\documentclass{article}

     \usepackage[preprint]{neurips_2022}

\usepackage[utf8]{inputenc} 
\usepackage[T1]{fontenc}    
\usepackage{hyperref}       
\usepackage{url}            
\usepackage{booktabs}       
\usepackage{amsfonts}       
\usepackage{nicefrac}       
\usepackage{microtype}      
\usepackage{xcolor}         

\hypersetup{colorlinks, linkcolor=blue, citecolor=blue, urlcolor=blue}

\usepackage{amsmath}
\usepackage{graphicx}
\usepackage{subcaption}

\newtheorem{remark}{Remark}

\title{Computing the ensemble spread from deterministic weather predictions using conditional generative adversarial networks}

\author{%
  Rüdiger Brecht \\
  Department of Mathematics \\
  University Bremen\\
  Bremen, Germany \\
  \texttt{rbrecht@uni-bremen.de}
   \And
   Alex Bihlo \\
   Department of Mathematics and Statistics \\
   Memorial University of Newfoundland\\
  St. John's, Canada  \\
  \texttt{abihlo@mun.ca}
}

\begin{document}

\maketitle

\begin{abstract}
Ensemble prediction systems are an invaluable tool for weather forecasting. Practically, ensemble predictions are obtained by running several perturbations of the deterministic control forecast. However, ensemble prediction is associated with a high computational cost and often involves statistical post-processing steps to improve its quality.
Here we propose to use deep-learning-based algorithms to learn the statistical properties of an ensemble prediction system, the ensemble spread, given only the deterministic control forecast. Thus, once trained, the costly ensemble prediction system will not be needed anymore to obtain future ensemble forecasts, and the statistical properties of the ensemble can be derived from a single deterministic forecast. We adapt the classical \texttt{pix2pix} architecture to a three-dimensional model and also experiment with a shared latent space encoder--decoder model, and train them against several years of operational (ensemble) weather forecasts for the 500 hPa geopotential height. The results demonstrate that the trained models indeed allow obtaining a highly accurate ensemble spread from the control forecast only. 
\end{abstract}

\section{Introduction}

To accurately predict weather and climate numerical weather prediction (NWP) models are being used~\cite{bauer2015quiet}. Recently, machine learning methods have received more attention as an alternative approach for weather and climate prediction. For example in \cite{bihlo2021generative,dueben2018challenges,weyn2021sub} neural networks are trained on reanalysis data to produce purely data-driven weather forecasts. While still in its infancy, if successful these data-driven approaches would enable issuing weather forecasts at several orders of magnitude faster than conventional NWP models. 

Owing to the inherent uncertainty in the atmospheric dynamical system~\cite{lorenz1963deterministic} it has been recognized early on in the development of NWP that a measure of uncertainty of a numerical weather forecast can substantially enhance the value of these forecasts. This gave rise to the field of ensemble weather prediction~\cite{leutbecher2008ensemble}, which aims to quantify the various sources of uncertainty in NWP models, chief of which are uncertainty in the initial conditions and errors in the numerical model formulation. To overcome these uncertainties, in addition to the single deterministic weather forecast, an ensemble of perturbed forecasts is generated whose overall divergence, or spread, ideally will provide a measure of the uncertainty in the deterministic prediction. The main limiting factor in generating the ensemble is still computational in nature as each ensemble run takes up computational resources thereby limiting the total number of such ensembles, typically less then 100, that can be computed on an operational basis. 

Here, we show that it is possible to use deep neural networks to directly learn the spread from a deterministic weather forecast, thereby possibly alleviating the need for costly numerical ensemble prediction models. We illustrate this idea by predicting the 500 hPa geopotential height, a measure of pressure in the middle of the troposphere. While still a rather stark simplification of the full three-dimensional structure of the atmosphere, prediction of the 500 hPa geopotential has been routinely considered as a first step in the history of weather forecasting, see the seminal work on NWP~\cite{char50a}, and more recently also for purely data-driven models~\cite{bihlo2021generative,dueben2018challenges,weyn2021sub}.

\section{Related work}

As other fields of science, also meteorology has seen an unprecedented rise in use of machine learning, and specifically deep learning, to areas that have traditionally been tackled using methods of numerical analysis and scientific computing. Deep learning has been investigated as a full replacement for NWP models in, e.g.~\cite{bihlo2021generative, dueben2018challenges, pathak2022fourcastnet, weyn2019can}. While still not competitive in comparison with traditional NWP at high resolution, the results obtained are often comparable at coarser resolution~\cite{pathak2022fourcastnet}. Deep learning has also been proposed as an alternative to subgrid-scale parameterization~\cite{gentine2018could}, it has been used extensively for precipitation nowcasting~\cite{shi2015convolutional}, downscaling of meteorological fields~\cite{bano2020configuration} and most recently also in combination with differential equations-based methodologies~\cite{bihlo2022physics}. 

Ensemble prediction methods using deep learning were investigated by~\cite{bihlo2021generative} who trained a conditional generative adversarial network to forecast the a variety of meteorological quantities, with the ensemble model being realize via Monte-Carlo dropout. In~\cite{scher2021ensemble} a variety of methods for generating ensemble predictions using deep learning were investigated, which included random initial perturbations, retraining of the neural networks, Monte-Carlo dropout, and ensembles based on singular value decomposition. In  \cite{gronquist2019predicting} it was investigated whether deep convolutional neural networks could be used to reduce the number of ensemble members while still maintaining the correct ensemble spread. Post-processing of computed ensemble members using neural networks was put forth in~\cite{gronquist2021deep}. In~\cite{weyn2021sub}, ensemble forecasting based on convolutional neural networks on a cubed sphere were proposed as a tool for sub-seasonal forecasting. Generating ensembles via stacking of sub-neural networks trained on a different subset of the predictive meteorological variables each was investigated in~\cite{clare2021combining}. Most closely related to our approach is the work of~\cite{scher2018predicting}, who trained a regression model based on a convolutional neural network to forecast model errors and ensemble spread, albeit as an integral measure over the entire forecast domain rather than as a gridpoint-by-gridpoint measure (via a paired video-to-video translation model) as considered here.

\section{Methods}

In this section we describe the data used to train the neural network and also the neural network architecture.

\subsection{Data} \label{sec:data}

As discussed in~\cite{scher2018predicting}, a main challenge in obtaining suitable training data for meteorological applications is that operational weather prediction centres routinely update their models, which will correspondingly have an impact on the statistical distribution of the training data. This is why almost all of the aforementioned applications used the ERA5 reanalysis dataset, see~\cite{era5}, rather than the archived operational weather prediction data. For ensemble forecasting this is unfortunately not practical as the ERA5 data archive does not store complete ensembles. 

Instead, here we use the operational ECMWF IFS data. We downloaded the data on a regular longitude--latitude grid with spatial resolution of $0.5^\circ\times 0.5^\circ$, corresponding to an orthogonal grid of $720\times 360$ grid points. We retrieved the ensemble data initialized at 00 UTC, for the next 96 hours at a temporal resolution of 6 hours corresponding to 16 time steps per ensemble run, from the years 2010--2020. We use the first 80\% of the data (corresponding roughly to the years 2010--2018) for training, the next 10\% of the data (corresponding roughly to the year 2019) for validation and the last 10\% of the data (corresponding roughly to the year 2020) for testing. The meteorological field considered here is the 500 hPa geopotential height, which is a measure of the pressure in the middle of the troposphere. The advantage of this predictor is that for an idealized barotropic atmosphere the evolution of this geopotential height is only dependent on the geopotential height itself rather than other atmospheric predictors such as temperature, the three-dimensional wind field and the specific humidity, see~\cite{holt12a}. This justifies to predict the 500 hPa geopotential ensemble spread given the 500 hPa geopotential height alone. While the real atmosphere is seldom purely barotropic, the first successful NWP was carried out for the 500 hPa geopotential height, see~\cite{char50a}. While we expect using a more complete description of the atmospheric state would improve the accuracy of the ensemble spread prediction presented here, for high-resolution atmospheric data this soon becomes a substantially more resource-intensive endeavour, which we reserve for future research.

\subsection{Neural network architecture} \label{sec:network}

We regard the problem of reconstructing the temporal evolution of the ensemble spread from the deterministic control forecast as a \textit{paired video-to-video translation} problem. This is an extension of the more classical paired image-to-image translation problem, for which a variety of methods based on convolutional neural networks can be used, with conditional generative adversarial network (GAN) approaches being among the most powerful options. 

\begin{figure}
\centering
    \includegraphics[clip, trim=0cm 0cm 0cm 0cm,width=\textwidth]{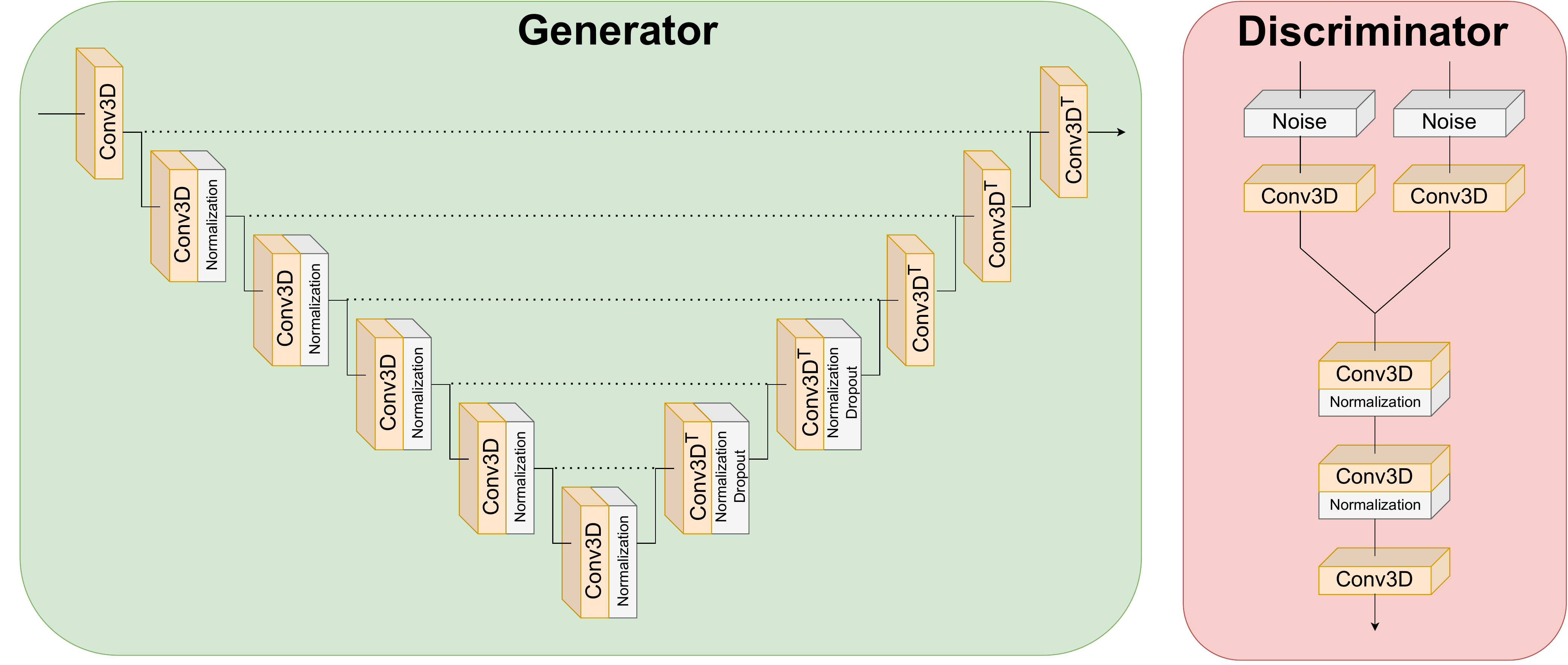}
    \caption{Neural network architecture for the generator and discriminator of the \texttt{pix2pix3D} model.}
	\label{fig:PixArchitecture}
\end{figure}

Specifically, we consider a model that is based on the classical \texttt{pix2pix} architecture adapted from \cite{isola2017image} for three-dimensional datasets. This architecture constitutes a conditional GAN and consists of a generator and a discriminator network, see Fig. \ref{fig:PixArchitecture}. 

The generator is a three-dimensional encoder--decoder U-net which consists of several three-dimensional convolutional layers to reduce the dimension of the input. Batch normalization is used after each convolutional layer. The output of each convolution layer is then concatenated to transposed convolutional layers which increase the latent dimension back to the original input/output shape. Additionally, a dropout rate of 50\% is added to the transposed convolutional layers. The precise model description can be found in Tables~\ref{tab:generator} and~\ref{tab:discriminator} in the appendix.

The discriminator takes an generator input output pair as the input and concatenates this input after applying a layer noise and a convolution. Then, several convolutions with batch normalization are applied to reduce the dimension. 

To train the neural network models, we use the geopotential as an input and train the model output to match the spread of the ensemble forecast. We implement the models in TensorFlow 2.8\footnote{Code will be made available at \url{https://github.com/RudigerBrecht/Computing_ensemble_spread_using_GANs}} and trained them on a machine with a dual NVIDIA RTX 8000 GPU. The training for 10 epochs for one model took about 5 hours.

\begin{remark}
An alternative to the three-dimensional \texttt{pix2pix} model proposed here would be any of the shared latent space models as pursued, e.g., in~\cite{huan18a,liu17a}. The main idea of this line of work is that both input and target images share a (possibly only partially) overlapping latent space, which requires the training of two autoencoders (for input and output image reconstruction) as well as a cross-domain model that transfers images from the input to the output domain, along with some discriminators to assess the quality of the produced results. The idea of a shared latent space is appealing since for the present problem this latent space may be interpreted as the common physics shared by the deterministic and ensemble predictions, with the respective input/output image domains being particular aspects of the shared physics of the atmosphere.

We have experimented with a variety of models from the shared latent space family but haven't found them to yield superior results to the rather simple three-dimensional \texttt{pix2pix} model. Thus, for the ease of implementation and speed of training we exclusively present results for this three-dimensional \texttt{pix2pix} model here.
\end{remark}



\subsection{Baseline models}

To compare our approach, we consider two meteorologically meaningful baselines. The first baseline is the \textit{climatological mean ensemble spread}, which is the ensemble spread for one particular day of the year averaged over the training dataset. We should like to note here that a true climatological mean is typically computed over a 30 year time frame, which unfortunately is not possible for the present case as we do not have a long enough training set. Thus, our ``climatology'' is based on seven years worth of data only. The second baseline is the \textit{persistence ensemble spread}, based on the premise ``The weather tomorrow is the weather today.'' The persistence ensemble spread is hence the true ensemble spread from the previous day, shifted to the current day. 

While it would be beneficial to consider other baseline models as well, notably those based on other machine learning methodologies, as indicated in the \textit{Related work} section, to the best of our knowledge this paper is the first that attempts to directly forecast the full gridpoint-to-gridpoint ensemble spread directly from the ECMWF IFS control forecast data via a paired video-to-video translation model.

\subsection{Forecast error norms}

To evaluate our models we use the root mean squared error (RMSE), the structural similarity index measure (SSIM)
and the ensemble spread integral 
:

\begin{align*}
	\text{RMSE}(x,y) &=\sqrt{\overline{(x-y)^2}},
	\\
	\text{SSIM}(x,y) &= \frac{(2\mu_x\mu_y+C_1)(2\sigma_{xy}+C_2)}{(\mu_x^2+\mu_y^2+C_1)(\sigma_{x}+\sigma_{y}+C_2)},
	\\
	\text{Spread Integral}(x) &= \frac{1}{|S|} \int_S x ~ds.
\end{align*}
Here, the overbar denotes spatial averaging, $\mu_x$ and $\mu_y$ denote the means of the images $x$ and $y$, $\sigma_{x}$ and $\sigma_{y}$ are the standard deviations of these images, and $\sigma_{xy}$ is the co-variance of these images; $C_1$ and $C_2$ are constants, see~\cite{wang04a} for further explanations on the SSIM. 
Here, $S$ denotes the surface of the Sphere and $ds$ is the area element in spherical coordinates.

The smaller the RMSE the better the result is and the closer the SSIM is to 1 the more similar the two images are.
Note that the ensemble spread integral is not a verification measure but we include it to evaluate if the generated spread of the models is over- or under-disperse, see Section~\ref{sec:results} for further discussions.

\section{Results}\label{sec:results}

In this section we compute the spread for the year 2020 using the baseline and \texttt{pix2pix3D} methods. Then, we also demonstrate how we can further improve the spread of the \texttt{pix2pix3D} model using various post-processing methods.

\subsection{Forecast spread evaluation} \label{sec:forecast}

We evaluate the spread generated from the different methods for a randomly selected day, 31st of March 2020. For other days the results are both qualitative and quantitative similar and available on this project's GitHub repository. To compare the spread against the ground truth we use the RMSE and SSIM.

\begin{figure}[!ht]
\centering
    \includegraphics[width=\textwidth]{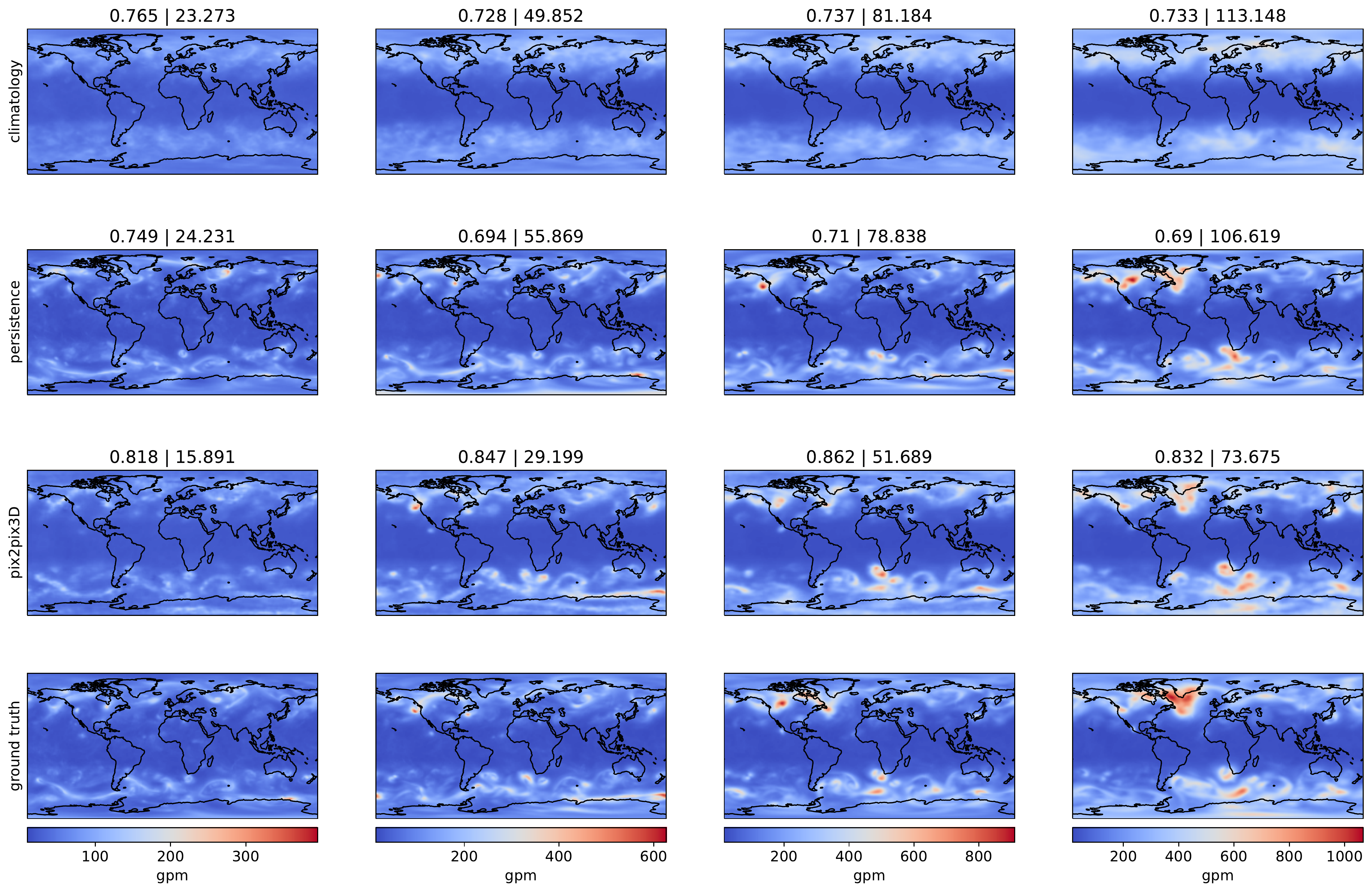}
    \caption{Ensemble spread forecast for 31st March 2020 at hours 18, 42, 66 and 90. The title shows the SSIM and RMSE. \textbf{Top row:} Climatological ensemble spread, \textbf{middle row:} persistence ensemble spread, \textbf{third row:} \texttt{pix2pix3D} ensemble spread and \textbf{bottom row:} ground truth.}
	\label{fig:spread_model}
\end{figure}

In Fig.~\ref{fig:spread_model} we present snapshots of the predicted spreads at different times. We observe that the climatological spread is rather blurry and only captures correctly the overall higher variability in the multitudes and the lower variability in the tropics, but 
does not capture the locations with locally higher spread for the particular given day. The persistence spread better captures these locations, however incorrectly predicts a high spread at locations where there is less uncertainty. Moreover, the spread shows a phase shift with the true spread at various locations, which is consistent with persistence spread being simply shifted in time by one forecast cycle. In contrast the \texttt{pix2pix3D} model predicts visually well the spread and also yields the lowest error norms, typically by a rather large margin. 

\begin{figure}[!ht]
\centering
    \includegraphics[width=\textwidth,trim={0cm 0cm 0cm 0cm},clip]{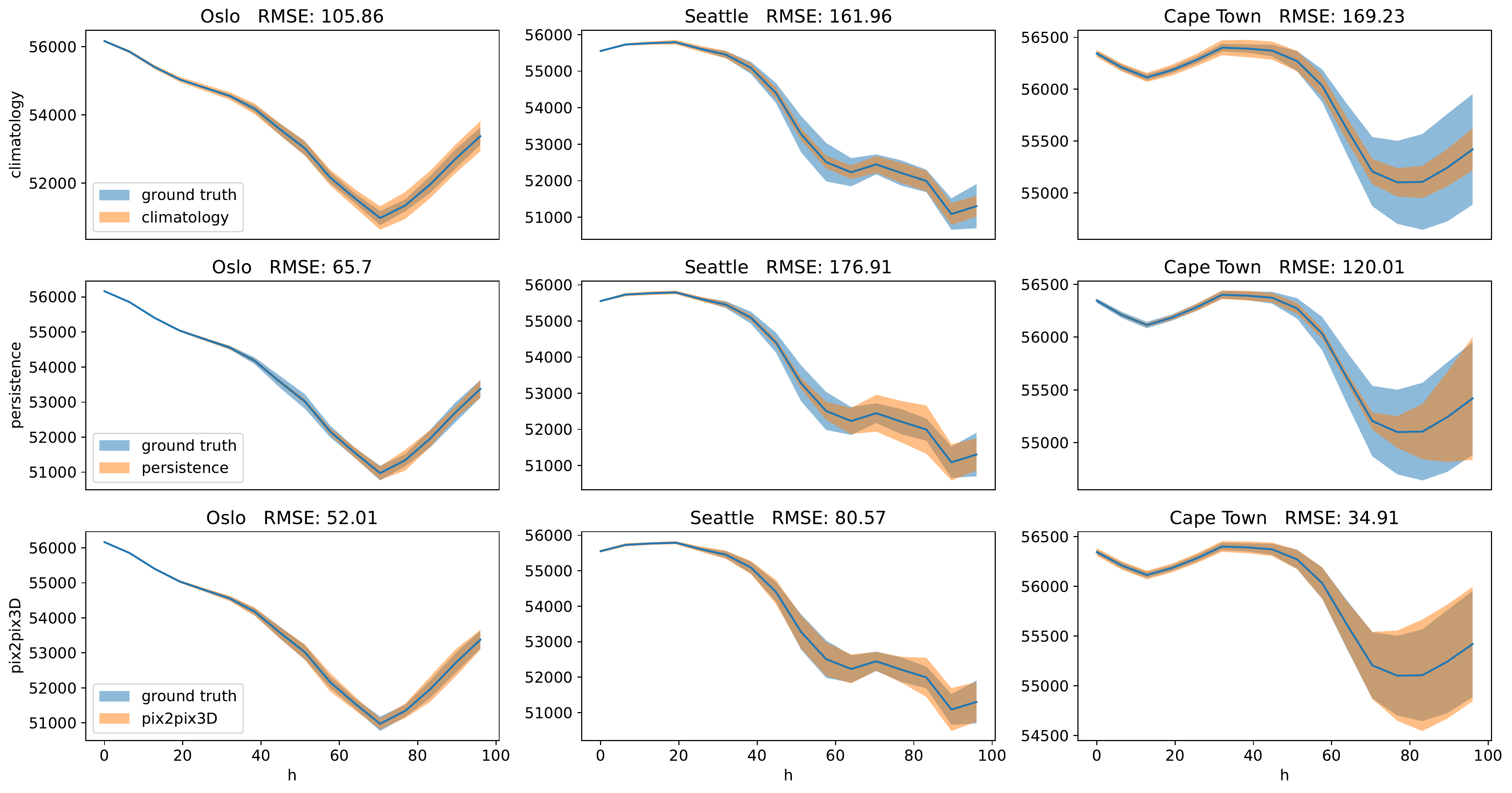}
    \caption{ Spread forecast for three different cities on 31st March 2020. Depicted is the control run plus and minus the respective ensemble spread.
    \textbf{Top row:} Climatological ensemble spread, \textbf{middle row:} persistence ensemble spread, \textbf{bottom row:} one realization of \texttt{pix2pix3D} ensemble spread.}	
	\label{fig:cities}
\end{figure}

In Fig.~\ref{fig:cities} we depict the spread as a function of time for three randomly selected cities. The results at other cities are both qualitatively and quantitatively comparable and not shown here due to page constraints. Both plots illustrate a good visual agreement between the true spread and the predicted spread using the \texttt{pix2pix3D} model, which strongly improves upon the baseline models.

For an overall verification we compute the RMSE and mean SSIM for the year 2020 for each forecast hour and show them as a function of time in Fig. \ref{fig:verification}. Initially the forecast uncertainty, and hence the spread, is small such that for the very first hours the SSIM of the climatological mean is higher than for the other methods. However, considering the RMSE we see that the other methods outperform this method, see Table \ref{table:1}. Moreover, we want to stress that the RMSE for the \texttt{pix2pix3D} model always gives better results compared to the baseline methods. 
Also in Fig. \ref{fig:verification} we analyze if the generated spread over- or under-estimates the true forecast uncertainty. We average for each forecast hour the mean spread over the test dataset. The persistence spread has the same magnitude as the ground truth spread, whereas the climatological and \texttt{pix2pix3D} spread are slightly higher. This is important as it is well known that the spread from NWP models (acting as our ground truth here) is under-dispersive, meaning it underestimates the true variability of the atmosphere~\cite{palm19a}.

\subsection{Forecast ensemble spread post-processing}\label{sec:forecastpost}

To improve the ensemble spread from a single uncalibrated ensemble prediction model we can use various methods of post-processing. For machine learning based models, 
we follow the ideas put forth by \cite{bihlo2021generative} and \cite{scher2021ensemble}, 
who proposed to train multiple versions of the same neural network architectures with different initial random weights and to use Monte-Carlo dropout. For the Monte-Carlo dropout ensemble we average 10 individual runs of a single \texttt{pix2pix3D} model. Also we train 5 instances of the same \texttt{pix2pix3D} model with different random initial weights and average their predictions.
Both methods yield and ensemble of ensemble spread prediction models which we evaluate in the sequel. Both methods will yield slight variations in the predicted ensemble spread for the same initial control forecast, with the goal to further improve the machine learning model based spread. Note that in contrast to conventional ensemble prediction systems, neural network based ensemble prediction systems are computationally inexpensive to run once trained, thereby allowing for more post-processing possibilities than are available in operational NWP today.

\begin{figure}
\centering
    \includegraphics[width=\textwidth,trim={0cm 0cm 0cm 0cm},clip]{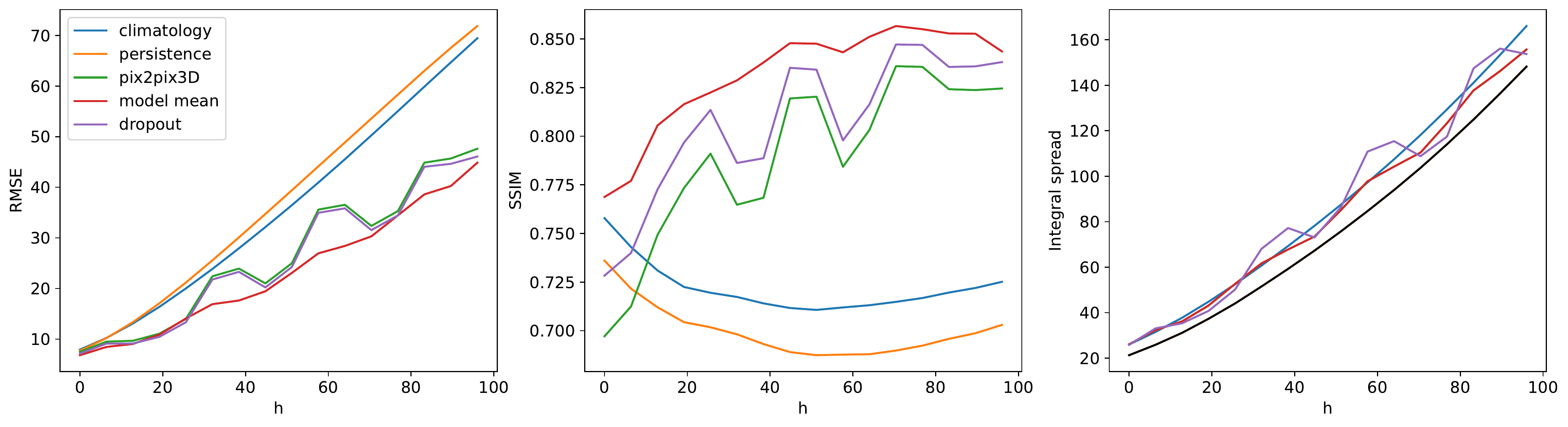}
    \caption{ Forecast verification metrics over the test dataset as a function of the forecast step.}
	\label{fig:verification}
\end{figure}

Figure ~\ref{fig:verification} and Table \ref{table:1} demonstrate that post processing the \texttt{pix2pix3D} model strongly improves upon all baseline models. Here, using the mean of multiple trained models yields in the most accurate ensemble spread. In particular, it smooths out the oscillatory behaviour of a single \texttt{pix2pix3D} model, yielding an overall lower (upper) bound for the RMSE (SSIM) of all models.

\begin{table}[]
\caption{Mean verification metrics over the test dataset.}
\begin{center}
\begin{tabular}{l||c|c}
        Method               &  RMSE & mean SSIM \\ \hline\hline
climatology                     &  51.747    &  0.722         \\ \hline
persistence                        &  53.184    &  0.700          \\ \hline
pix2pix3D                        &  39.850    &   0.789        \\ \hline
pix2pix3D dropout mean            &  38.773    &   0.807        \\ \hline
pix2pix3D model mean             &  \textbf{35.075}    &  \textbf{0.831}
\end{tabular}
\end{center}
\label{table:1}
\end{table}

\section{Conclusions}\label{sec:conclusion}

The present paper is devoted to the use of deep learning for ensemble weather forecasting. Ensemble forecasts are typically obtained by re-running a version of the operational NWP model with slightly perturbed initial conditions and model parameters. This renders obtaining ensemble forecasts computationally rather expensive as running an NWP model multiple times is costly and computationally burdensome, in particular for operational weather forecasting which is a time-sensitive endeavour. Hence, the number of ensemble members that can be computed is severely limited and typically of the order of 50 or less, which is several orders of magnitude smaller compared to the several millions of degrees of freedom in a state-of-the-art NWP model.

Here we investigated the possibility of obtaining the ensemble spread directly from the deterministic control forecast by interpreting the ensemble prediction problem as a paired video-to-video translation problem. We trained a conditional GAN, an adaptation of the \texttt{pix2pix} model, for this purpose. This model succeeds for this task as evidenced by considerably improving upon two important meteorological baselines, the climatological and the persistence spreads. Further improvements are observed with post-processing using Monte-Carlo dropout and retraining the network multiple times and computing the mean ensemble spread over all models.

Interestingly, we found that our models tend to produce slightly larger spreads than the ground truth spread. This is relevant since it is well-known that model spreads tend to underestimate the true variability of the atmosphere, i.e.\ they are under-dispersive~\cite{palm19a}. This would suggest that the spread obtained from our machine learning based ensemble prediction systems has the potential to better capture the true atmospheric variability, which we aim to investigate in a future study.

A limiting factor of the present study is that ensemble predictions are operationally issued only once or twice per day and thus training data is rather scarce, in particular compared to reanalysis data on which meteorological deep learning models are traditionally trained on. The results of the present study might warrant operational NWP centres to consider creating a research archive of ensemble predictions ran at a higher frequency so as to provide a larger database of training data, which would enable training larger and more sophisticated models as the one proposed here.

\section*{Acknowledgements}

The authors are grateful to Jason Brownlee and the developers of TensorFlow for providing codes for the \texttt{pix2pix} models adapted in this study.
This research was undertaken thanks to funding from the Canada Research Chairs program and the NSERC Discovery Grant program. This study is also supported by the Deutsche Forschungsgemeinschaft (DFG, German Research Foundation) – Project-ID 274762653 – TRR 181. The authors also acknowledge support from the ECMWF special project \textit{Mining 5th generation reanalysis data for changes in the global energy cycle and for estimation of forecast uncertainty growth with generative adversarial networks}.

\appendix

\section{Implementation details} \label{sec:impdet}

The generator and discriminator implementation details are given in Tables~\ref{tab:generator} and~\ref{tab:discriminator}.

\begin{table}[!ht]
\caption{Implementation details of the generator neural network.}
\begin{center}
\begin{tabular}{l||c|c|c}
Generator            & no.\ of filter & kernel size & strides \\ \hline \hline
convolution & 16     & (4,4,4)       & (2,2,2)  \\
convolution & 32     & (4,4,4)       & (1,2,2)  \\
convolution & 64     & (4,4,4)       & (1,2,2)  \\
convolution & 128    & (3,4,4)       & (1,2,2)  \\
convolution & 128    & (3,4,4)       & (1,2,2)  \\
convolution & 256    & (3,4,4)       & (1,2,2)  \\
convolution & 256    & (3,4,4)       & (1,1,1) \\ \hline
deconvolution & 256     & (3,4,4)       & (1,1,1)  \\
deconvolution & 128     & (3,4,4)       & (1,2,2)  \\
deconvolution & 128     & (3,4,4)       & (1,2,2)  \\
deconvolution & 64    & (4,4,4)       & (1,2,2)  \\
deconvolution & 32    & (4,4,4)       & (1,2,2)  \\
deconvolution & 16    & (4,4,4)       & (1,2,2)  \\
deconvolution & 16    & (4,4,4)       & (2,1,1) \\ 
deconvolution & 16    & (4,4,4)       & (2,1,1) \\ 
\end{tabular}
\end{center}
\label{tab:generator}
\end{table}

\begin{table}[!ht]
\caption{Implementation details of the discriminator neural network.}
\begin{center}
\begin{tabular}{l||c|c|c}
Discriminator            & no.\ of filter & kernel size & strides \\ \hline \hline
convolution & 32     & (4,4,4)       & (1,2,2)  \\
convolution & 256     & (4,4,4)       & (2,2,2)  \\
convolution & 1     & (4,4,4)       & (2,1,1)  \\
\end{tabular}
\end{center}
\label{tab:discriminator}
\end{table}

\bibliography{mybib}

\end{document}